\documentclass[a4paper,conference]{IEEEtran}
\usepackage{graphicx}
\usepackage{comment}
\usepackage{amsmath,amssymb} 
\usepackage{color}
\usepackage{algorithm}
\usepackage{algpseudocode}
\usepackage{subcaption}
\usepackage{multirow}
\usepackage{caption}
\usepackage{amsfonts}
\usepackage{wrapfig}
\usepackage[utf8]{inputenc}

\ifCLASSINFOpdf
\else
\fi
\begin{document}
%
\title{Quaternion Capsule Networks}

\author{\IEEEauthorblockN{Bar{\i}\c{s} \"{O}zcan}
\IEEEauthorblockA{Department of Computer Science \\
\"{O}zye\u{g}in University \\
\.{I}stanbul, Turkey \\
Email: baris.ozcan.10097@ozu.edu.tr}
\and
\IEEEauthorblockN{Furkan K{\i}nl{\i}}
\IEEEauthorblockA{Department of Computer Science \\
\"{O}zye\u{g}in University \\
\.{I}stanbul, Turkey \\
Email: furkan.kinli@ozyegin.edu.tr}
\and
\IEEEauthorblockN{Furkan K{\i}ra\c{c}}
\IEEEauthorblockA{Department of Computer Science \\
\"{O}zye\u{g}in University \\
\.{I}stanbul, Turkey \\
Email: furkan.kirac@ozyegin.edu.tr}}


%


\maketitle

\begin{abstract}
Capsules are grouping of neurons that allow to represent sophisticated information of a visual entity such as pose and features. In the view of this property, Capsule Networks outperform CNNs in challenging tasks like object recognition in unseen viewpoints, and this is achieved by learning the transformations between the object and its parts with the help of high dimensional representation of pose information. In this paper, we present Quaternion Capsules (QCN) where pose information of capsules and their transformations are represented by quaternions. Quaternions are immune to the gimbal lock, have straightforward regularization of the rotation representation for capsules, and require less number of parameters than matrices. The experimental results show that QCNs generalize better to novel viewpoints with fewer parameters, and also achieve on-par or better performances with the state-of-the-art Capsule architectures on well-known benchmarking datasets. Our code is available\footnote{https://github.com/Boazrciasn/Quaternion-Capsule-Networks.git}.

\end{abstract}


%
\IEEEpeerreviewmaketitle

\section{Introduction}
With the rapid rise of CNNs in the recent years, unprecedented performance improvements have been achieved in various Computer Vision tasks (e.g. object recognition, image segmentation and retrieval). Since AlexNet \cite{krizhevsky2012imagenet} made a drastic performance improvement on ImageNet challenge, the efforts to solve these problems led to the discovery of deeper and more complex networks that require larger datasets \cite{ren2015faster,he2016residual,Redmon_2016_YOLO,DeeperConv}. However, human level object recognition abilities are still yet to be achieved in terms of generalization to viewpoints as CNNs fail to generalize to geometric variations. Since the structural relationships between the entities are not modelled by CNNs, it requires many samples for a specific object from different viewpoints.

First generation Capsule architectures \cite{DynRouteCaps,hinton2018matrix} address these deficiencies of CNNs by exploiting the linear relationship between the poses of an object and its parts. However, this mechanism requires a group of neurons, called \textit{capsules}, with high dimensional tensor outputs in each layer, as distinct from traditional NNs. This allows to formulate more sophisticated relationships between capsules to learn the aforementioned linear relationship that provides viewpoint invariant relations. In the context of an object, connected capsules have a part-whole relationship where child capsules in lower level represents the parts while a parent capsule in higher level modelled by the routing mechanism is considered as the whole. In principle, routing mechanisms cluster the transformations of lower level capsule outputs to calculate the corresponding higher level capsule output where these correspondences can be fully-connected or convolutional, etc.  Even though Capsule Networks outperform the state-of-the-art in small datasets (\textit{e.g} MNIST \cite{DynRouteCaps}, smallNORB \cite{hinton2018matrix,VariationalCaps}, there is a limited number of studies on larger datasets in the literature due to the computational complexity of transformations and the routing mechanisms. The previous capsule architectures aimed to learn linear manifold between the object and a $4\times4$ pose matrix or a feature vector of its parts. Particularly, the feature vector of an object and its part does not guarantee a linear manifold of the transformation space. In case of using a pose matrix, to guarantee the linear manifold between capsules, it should be assumed that connected capsules only learn the rotations in-between higher and lower level parts.

\begin{figure}[t]
\begin{center}
\includegraphics[width=1\linewidth]{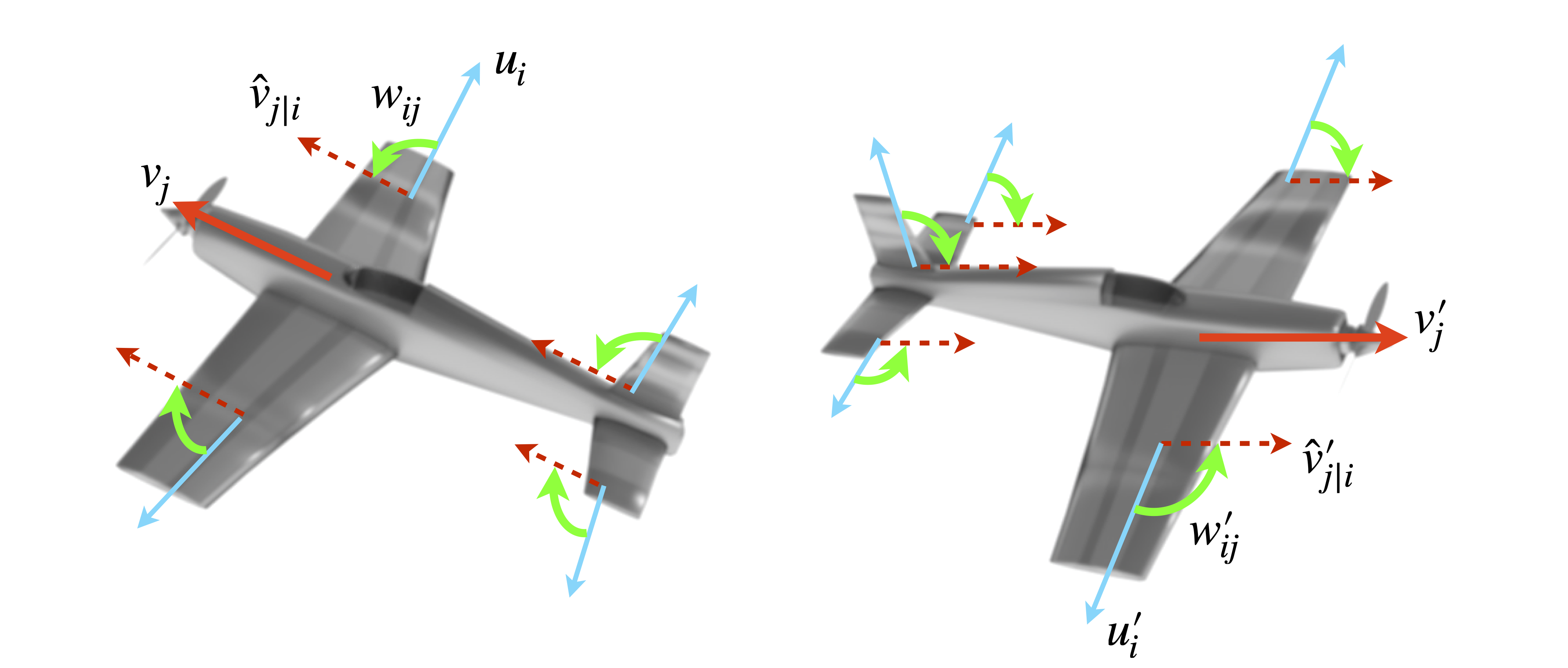}
\end{center}
\caption{An illustration of rotations between child capsules and their parent object capsule. For both orientations of the plane, the rotations between child capsules and the parent capsule should stay the same ($w_{ij} = w_{ij}'$). \textbf{Blue:} Child capsule pose, \textbf{Red:} Parent capsule pose, \textbf{Dashed Red:} Votes for parent capsule, \textbf{Green:} Intrinsic rotation between child and parent poses.}
\label{fig:fenerucak}

\end{figure}

From the viewpoint of a visual entity, all its parts should agree on the orientation and the origin of the coordinate system of this entity. In case of convolutional connections, the agreement on the origin can be discarded with the assumption of the origin of a visual entity and all its parts are located at the center of the same receptive field, and hence the agreement on the origin is safe to ignore. When the visual entity does not have any deformable parts, there is a fixed mapping from each part to the parent entity in the form of rotation, as illustrated in Figure \ref{fig:fenerucak}. Moreover, convolutionally-connected Capsule Networks that only represent the orientation and its learned transformation set, which is restricted to the rotations, can still generalize to novel viewpoints. In this way, the computational burden and the size of the network are significantly reduced as the network seeks for a solution in a smaller space in case of transformations. In parallel with this intuition, Stacked Capsule Autoencoders (SCAE) \cite{kosiorek2019stacked} represent the relationships in 3D space with $3\times3$ matrices.

In case of the representation of the rotations, using quaternions have several advantages compared to the rotation matrix. First, quaternions do not suffer from gimbal lock \cite{grassia1998practical,pavllo2018quaternet}, as distinct from rotation matrices, Euler angles, and exponential maps. Second, rotation matrices must be orthogonal, and quaternions must be a unit vector for a proper representation of the rotation. However, neither can guarantee to keep this property after the weight updates. At this point, it is harder to ensure the orthogonality in the rotation matrix, and better to normalize quaternions. Additionally, in the previous studies, it is demonstrated that quaternions are successful in restoring the spatial relations \cite{matsui2004quaternion}, and can be extracted from images \cite{trabelsi2017deepComplex, gaudet2018deepQuat}. Lastly, using quaternions reduces the number of parameters for each 3D rotation to be learned from $9$ to $4$.

By incorporating the aforementioned properties of quaternions and capsules, we propose Quaternion Capsules, a novel form of capsules where the capsule orientations and their rotations are governed by quaternion algebra. Quaternion Capsules have three main contributions to both Capsule Network and Quaternion Network literature. First, capsule representation with quaternions is proposed, and mappings between them in quaternion algebra is formulated. Secondly, the rotation axis is a constant in the previous studies of feed-forward Quaternion Networks. However, Quaternion Capsule Networks (QCNs) have the ability to learn the rotation axis along with the rotation angle, in a similar way to Recurrent Quaternion Neural Networks (QRNN) \cite{parcollet2018quaternionRNN}. Lastly, extracting pose and activations from the input image is divided into two spatially-aligned branches that allow to use different sub-networks with different complexities for different modalities. Experimental results show that Quaternion Capsules achieve the state-of-the-art performance in case of generalization to novel viewpoints. QCN also attains on-par performance with Matrix Capsules on well-known benchmarking datasets with the only half of the parameters without any hyper-parameter tuning or architecture changes with respect to the dataset.

\section{Related Work}
\textbf{Capsule Networks:} Capsules are basically encapsulation of multiple neurons that allows to represent pose information of visual entities. This idea is first introduced in Transforming Auto-Encoders \cite{hinton2011transforming}, which requires a set of previously defined transformation matrices in order to extract this information. Dynamic routing between Capsules \cite{DynRouteCaps} and Matrix Capsules with EM Routing \cite{hinton2018matrix} can learn these transformations via back-propagation. The general approach in all Capsule Networks are analogous to parse trees, where small entities (children) in the image are combined to generate a representation of a larger and more complex entity (parent). The process of constructing the parent object from its children is called \textit{routing}. Different routing algorithms (\textit{i.e.} Dynamic Routing \cite{DynRouteCaps} and EM Routing \cite{hinton2018matrix}) between capsule layers are proposed in the literature. The first is a routing mechanism that capsule poses are represented as vectors and their magnitudes are considered as the existence probabilities (activation). It has been proven that fully-connected capsules with Dynamic routing is highly efficient in segmenting overlapping digits, and it achieves the state-of-the-art performance on MNIST with less number of parameters. The latter is a more advanced routing algorithm that represents pose information with matrices, and estimates capsule activations with the help of Expectation-Maximization algorithm. In this design, capsule layers are convolutionally-connected, and capsule activations are independent from their pose information. This form of capsules outperforms the state-of-the-art CNN-based architectures by a significant margin on smallNORB \cite{lecun2004learning} dataset by reducing the error by 45\% and making the network more robust to the white-box adversarial attacks \cite{kurakin2016adversarial,brendel2017comment}, when compared to CNNs.

In addition to these fundamental studies in Capsule Network literature, a recent routing algorithm that splits the routing mechanism into two branches are proposed in Neural Network Encapsulation \cite{Li2018NeuralNE} to decrease the complexity of the routing process. Group Equivariant Capsule Networks \cite{lenssen2018groupCaps}, on the other hand, merges Group CNNs \cite{cohen2016group} with the Capsule idea in order to guarantee equivariance. Most recently, Variational Bayesian learning for capsules are introduced \cite{VariationalCaps} with the achievement of the state-of-the-art performance on smallNORB dataset. Moreover, learning the parts of an object from unlabelled data is a fundamental problem for all previous Capsule Network variations, and this problem is addressed by SCAE \cite{kosiorek2019stacked} with the help of a two-staged capsule auto-encoder that detects the parts, and combines these parts to form the objects. Lastly, the performance of Capsule Network variants is observed on different image-related tasks such as medical image segmentation \cite{duarte2018VideoCapsNet}, video action detection \cite{lalonde2018SegCaps} and fashion image retrieval \cite{kinli2019RCCapsNet} by modifying the main capsule structure slightly according to a particular task. All aforementioned studies focus on the routing mechanism, learning capsule representations in an efficient manner, or experimenting capsules on different tasks. In this study, we disparately focus on representing pose information and its transformations in a more compact way, with the help of quaternion algebra.

\textbf{Quaternion Networks:} Complex-valued neural networks have been an active research field in the last decades \cite{hirose2004complex,zimmermann2011comparison}, but with a limited influence until the applications on RNNs proving that complex-valued RNNs learn faster \cite{danihelka2016associative} and avoid vanishing gradients \cite{arjovsky2016unitary}. Moreover, the earlier studies show that complex-valued neural networks are easier to optimize \cite{nitta2002critical}, and have better generalization characteristics \cite{hirose2012complexgeneralization}. One of the earliest studies of complex numbers in feed-forward NNs is proposed by Trabelsi \textit{et al.} \cite{trabelsi2017deepComplex}, namely Deep Complex Networks. Following up to this study, Deep Quaternion Network was proposed by defining quaternion convolution operation, but not utilizing quaternion rotations \cite{gaudet2018deepQuat}. On the other hand, another variant of Quaternion Convolutional Networks \cite{zhu2018quaternionconv} represents each pixel in a colored image with a quaternion by considering RGB channels as imaginary parts. In this study, quaternion rotations are utilized with a constant rotation axis of $\sqrt{3}\left[\begin{matrix}i & j & k \end{matrix}\right]$, and thus only the rotation angles are learned during training. In case of RNNs, it is demonstrated that QRNNs reduce the number of parameters by a wide margin on speech recognition tasks while maintaining the test accuracy \cite{parcollet2018quaternionRNN}, \cite{parcollet2018quaternionEndtoEnd}. From Computer Graphics perspective, quaternions are widely used for rotations due to their useful properties such as smooth rotation and computational efficiency \cite{pletincks1988useQuat}. Moreover, quaternions are closely related to the Capsule idea in the context of the part-whole relationship since an object can be represented by the part-whole transformations in Computer Graphics. 

Our proposed architecture combines aforementioned concepts and Capsule Networks by forming capsules as quaternions, and it has several advantages. First, the rotations with quaternions can be represented with only 4 parameters, and it reduces the required parameters in the network to represent pose and transformations, when compared to $3\times3$ matrix in SCAE or $4\times4$ matrix in Matrix Capsules. Moreover, due to the rotation formulation of quaternions, the network is implicitly regularized by using quaternions. In quaternion rotation, rotor quaternions, which are learnable parameters in Quaternion Capsule Networks, must be unit quaternions. Therefore, a simpler version of weight normalization \cite{salimans2016weight} must be applied to the weights in order to ensure proper representation of the rotation. As a side effect of its computationally efficient nature as well as the immunity to ambiguity and gimbal lock, quaternions are better suited for optimization, in contrast to Euler angles. Lastly, the ability of directly learning the rotation angles by the network may open up an opportunity for better interpretation of the relationship between capsules in the future.


\section{Methodology}

Quaternion Capsule Networks are governed by the quaternion algebra, and the transformations are made via quaternion rotation. Note that EM Routing is applicable to Quaternion Capsules, and it allows us to directly use it for our proposed method. Quaternion Capsules use the exact same loss function and routing procedure as in \cite{hinton2018matrix}, so it helps us to make a fair comparison to Matrix Capsules. In this section, a general capsule notation is established to avoid any confusion with quaternion notation, then Quaternion Capsules are formulated, and lastly our proposed network architecture is explained in detail. 

\subsection{Capsule Notation}

In order to avoid any confusion with generic capsule notation inherited from \cite{Li2018NeuralNE}, for two consecutive capsule layers in layers $L$ and $L+1$, the outputs are
$\{ \left[u_{ i }, a_{ i } \right]\in \mathbb{R}^{ d_{ L } }\} ^{ n_{ L } }_{ i=1 }$ and
$\{ \left[v_{ j }, a_{ j } \right]\in \mathbb{R}^{ d_{ L+1 } }\} ^{ n_{ L +1} }_{ j=1 }$, respectively. While $u_i$ and $v_j$ represent pose, $a_i$ and $a_j$ represent activations of $i^{th}$ and $j^{th}$ capsules in layers $L$ and $L+1$. $n_L$ and $n_{L+1}$ are the number of capsules in layers $L$ and $L+1$, and $d$ is the dimensionality of the outputs. Capsule outputs in lower layers are projected to the higher level capsules by transforming their outputs that can be interpreted as a vote for the higher level capsule pose. The vote $\hat{v}_{j|i}$ from $i^{th}$ capsule in layer $L$ to $j^{th}$ capsule in layer $L+1$ is calculated by $\hat{v}_{j|i} = f_{ij}( u_i )$ where $f_{ij}(.)$ is a transformation operator. In \cite{hinton2018matrix}, $f_{ij}(.)$ is an affine transformation, which is defined as $f_{ij}( u_i ) = W_{ij} u_i$ where $W_{ij}\in \mathbb{R}^{ d_{L} \times d_{L+1}}$ is the transformation matrix between capsules. The output of higher level capsules $\left[v_j, a_j \right]$ is calculated by a routing process applied on the votes from the lower-level capsules. In \cite{DynRouteCaps}, activations for capsules are assumed to be the magnitude of the pose configuration represented by vectors, instead of matrices. In Quaternion Capsule Networks, the pose configuration is represented by quaternions, and $f_{ij}(.)$ is defined as quaternion rotation, instead of affine transformation.

\subsection{Quaternions}

 Quaternions are first proposed by Hamilton in 1883 as an extension of the complex numbers that represents an efficient way of computing rotations, when compared to Euler rotation matrix. A quaternion $q$ is a hyper-complex number that can be considered as a 4-dimensional vector, but governed by different algebraic rules than the standard vectors. Quaternions are defined as in \eqref{eqn:QuatNotation}, where $i, j, k$ denotes the imaginary axes.
 
 \begin{equation}
     \label{eqn:QuatNotation}
     q = q_0 + q_1 i + q_2 j + q_3 k, \quad q_0, q_1, q_2, q_3 \in \mathbb{R}
 \end{equation}
 
 Quaternions are often divided into two parts, a scalar part $s$ and a vector part $\mathbf{v}$ as given in \eqref{eqn:QuatVec}, to be able to define algebraic operations in a compact form. 

 \begin{equation}
     \label{eqn:QuatVec}
     q = \left[ \begin{matrix}s_q,  & \mathbf{v_q} \end{matrix} \right]
 \end{equation}
 
 The governing algebraic rules over quaternions are as follows:
 
\begin{equation}
    \label{eqn:QuatArith1}
    i^2=j^2=k^2=ijk=-1
\end{equation}

\begin{equation}
    \label{eqn:QuatArith2}
    ij=k, jk=i, ki=j, ji=-k, kj=-i, ik=-j
\end{equation}
\vspace{1mm}

To formulate quaternion rotation, quaternion conjugate and quaternion product must be defined for two quaternions $q$ and $p$.
 
 \begin{equation}
     \label{eqn:quats}
     q = \left[ \begin{matrix}s_q,  & \mathbf{v_q} \end{matrix} \right], \quad 
     p = \left[ \begin{matrix}s_p,  & \mathbf{v_p} \end{matrix} \right]
 \end{equation}
 \vspace{1mm}
 
 \textbf{Quaternion conjugate:}
 \begin{equation}
     \label{eqn:conjugate}
     q^{*} = \left[ \begin{matrix}  s_q, & -\mathbf{v_q}   \end{matrix} \right]
 \end{equation}
 \vspace{1mm}
 
\begin{figure}[t]
\begin{center}
\includegraphics[width=0.885\linewidth]{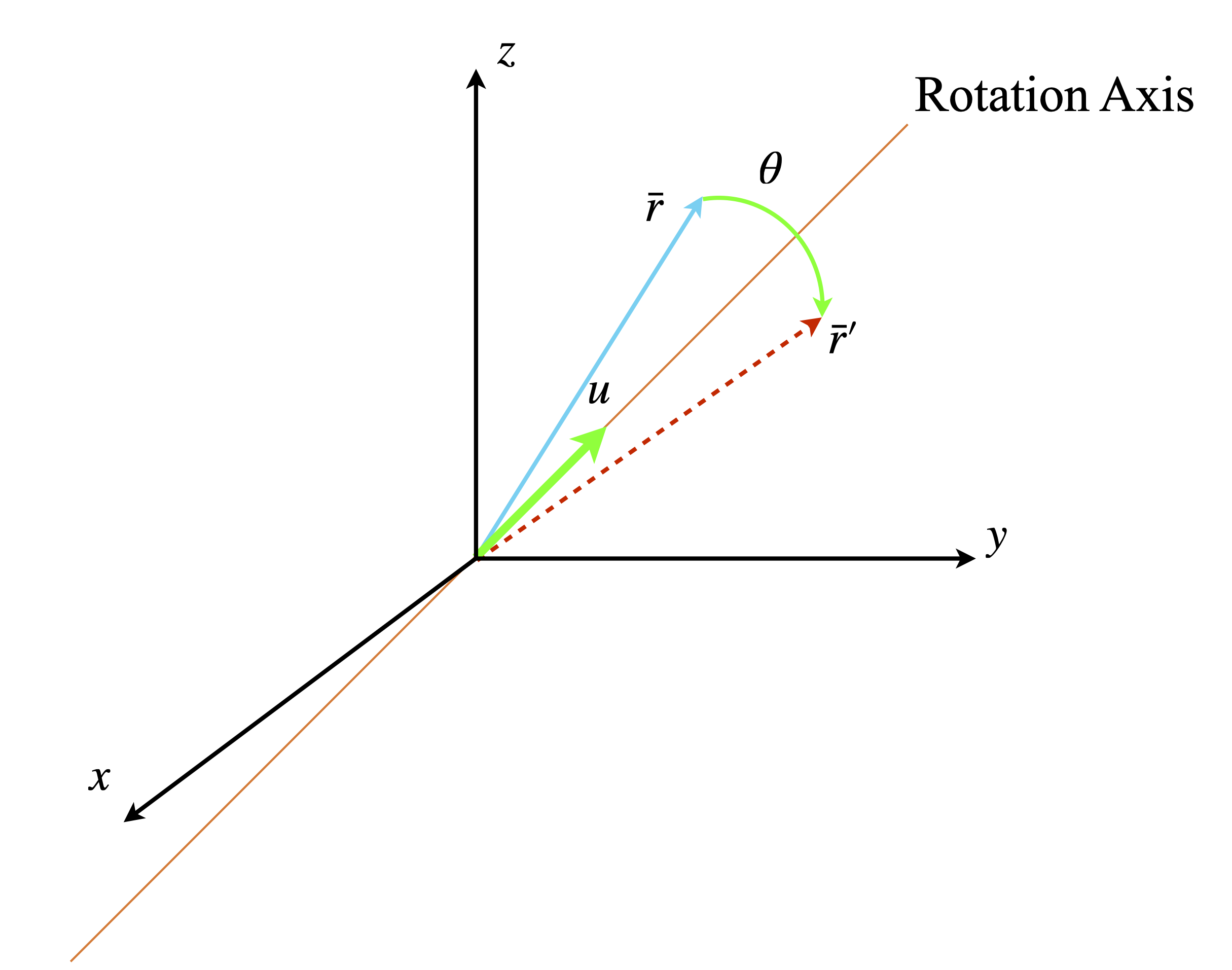}
\end{center}
\caption{Illustration of Equation \eqref{eqn:QuatRot}. Rotation of a pure quaternion $\Bar{r}$ by $\theta$ along the rotation axis $\mathbf{u}$.}
\label{fig:donek}
\end{figure}

 \textbf{Quaternion product:}
 \begin{equation}
          \label{eqn:product}
     q * p = \begin{bmatrix} s_q s_p - <\mathbf{v_q},\mathbf{v_p}> \\ 
                            s_q\mathbf{v_p} + s_p\mathbf{v_q} + \mathbf{v_q} \times \mathbf{v_p} 
             \end{bmatrix}^T
 \end{equation}
 
 \noindent
 where $<\mathbf{v_q},\mathbf{v_p}>$ and $\mathbf{v_q} \times \mathbf{v_p}$ are the dot product and the cross product between vectors $\mathbf{v_q}$ and $\mathbf{v_p}$, respectively. The rotation of a quaternion $\Bar{r}$ by $ \frac { \theta }{ 2 } $ around the axis $\mathbf{u}$ is represented in \eqref{eqn:QuatRot} (see Figure \ref{fig:donek}), where $q$ is a unit vector in the form of \eqref{eqn:q}, $q^{*}$ is the quaternion conjugate and $\Bar{r}$ is a pure quaternion whose scalar part equals to 0, as given in \eqref{eqn:r}.
 
 \begin{equation}
     \label{eqn:QuatRot}
     \Bar{r}' = q*\Bar{r}*q^{*}
 \end{equation}
 
 \begin{equation}
     \label{eqn:q}
     q = \left[ \begin{matrix} \cos{\theta},  & \mathbf{u} \sin{\theta} \end{matrix} \right]
 \end{equation}
 
 \begin{equation}
     \label{eqn:r}
     \Bar{r} = \left[ \begin{matrix}  0, & r  \end{matrix} \right]
 \end{equation}
 
 For a valid quaternion rotation, where the rotation is by $\frac { \theta }{ 2 }$ and the norm of $\Bar{r}$ is preserved, $q$ must be a unit quaternion and $\Bar{r}$ must be a pure quaternion. Therefore, using quaternion rotation, instead of transformations in capsules, implicitly regularize the network weights by constraining them to satisfy unit quaternion property.

 \subsection{Quaternion Capsules}
 A quaternion capsule in layer $L$ is defined as $\left[u_i, a_i \right]$ where $u_i$ is a pure quaternion and $a_i$ is the activation. The votes for parent capsules are calculated by quaternion rotation operation in \eqref{eqn:QuatRot}. As aforementioned before, EM Routing can be applied on the votes without any modification. To be consistent with the capsule notation, each capsule is denoted as $u$, and the parameters that rotate capsules and map capsule outputs in lower level to capsules in higher level layer are denoted as $w$. Thus, the vote from $i^{th}$ capsule in layer $L$ to $j^{th}$ capsule in layer $L+1$ is calculated as follows
 \begin{equation}
     \label{eqn:QuatVote}
    \hat{v}_{j|i} = f_ij(u_i) = w_{ij}*u_i*w_{ij}^{*}
 \end{equation}
 where capsule output is a pure quaternion $u_i = \left[ \begin{matrix}  0, & \mathbf{\Bar{u}_i}  \end{matrix} \right]$. To comply with the constraint of quaternion rotation, $\Bar{w}_{ij}$ must be a unit quaternion, therefore is defined as given in \eqref{eqn:NormalQ}.
 
 \begin{equation}
     \label{eqn:NormalQ}
     w_{ij} = \left[ \begin{matrix}  \cos{\theta_{ij}}, & \sin{\theta_{ij}}\mathbf{\frac{\Bar{w}_{ij}}{\left\| \Bar{w}_{ij} \right\|}}  \end{matrix} \right]
 \end{equation}
where $\mathbf{\Bar{w}} = \left[ \begin{matrix} w_1 i & w_2 j & w_3 k \end{matrix} \right]$ which refers to a similar weight normalization procedure as in \cite{salimans2016weight} by re-parameterizing the weights where the scale of the weight vector is set to 1 to ensure that the weight vector is a unit vector in the rotor quaternion. Moreover, $\theta_w$, $w_1$, $w_2$ and $w_3$ for each capsule are learned via backpropagation during training. Since the division of $\theta$ by $2$ in the original quaternion formulation is a constant, it is discarded on the weight formulation. Note that Quaternion Capsules can learn the rotation axis ${\Bar{w}}_{ij}$, in addition to the rotation angle $\theta_{ij}$, which is a novel feature for Quaternion Networks. Finally, once the votes are obtained, the parent capsules $\left[v_j, a_j\right]$ are calculated by EM Routing. 
 
 \begin{figure*}[t]
    \begin{center}
        \includegraphics[width=0.925\linewidth]{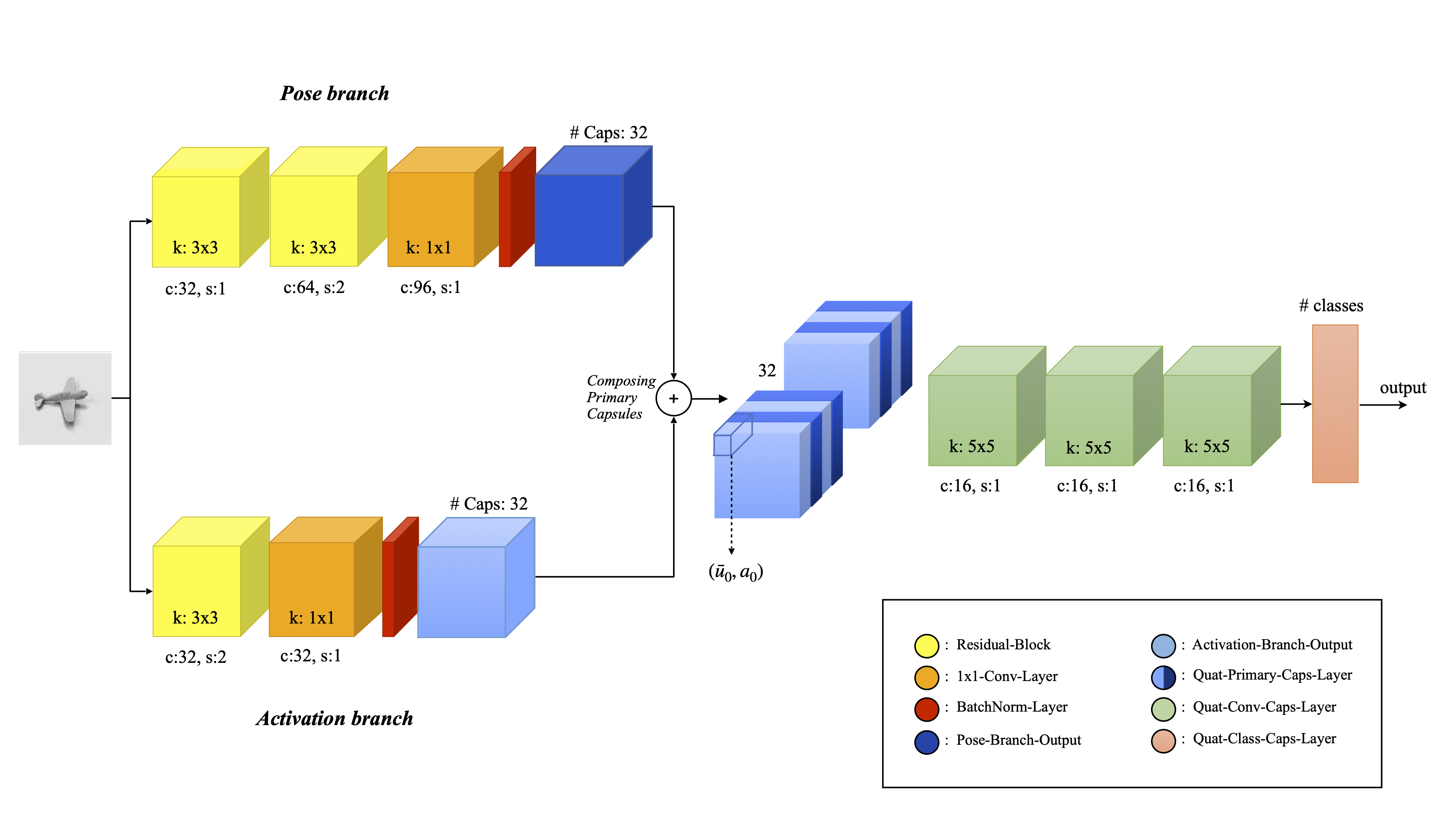}
    \end{center}
    \caption{Overview of QCN architecture. Input image is processed on Pose and Activation branches separately later to be merged to compose Primary capsules. Starting from Primary capsules, the network consists of three convolutional and one fully-connected capsule layers.}
    \label{fig:architecture}
\end{figure*}
 
\subsection{Architecture}
 
 In the literature, the activations and pose information for capsules are extracted by the same network despite the fact that the pose of a visual entity is independent from its existence probability. Our architecture inherently differs from other Capsule Networks in this manner. Instead of using two convolutional layers as in the original architecture \cite{hinton2018matrix}, pose $u_i$ and activations $a_i$ are extracted with the help of isolated branches that output spatially-aligned activations and pose for each capsule, as illustrated in Figure \ref{fig:architecture}. This design allows us to independently specify the size, or even the type of the network for pose and activations before constructing Primary capsules. In our design, a relatively deeper network is used in the pose branch since extracting pose information may be assumed to be a more complex problem than extracting the activations. Spatial alignment of the branches are ensured by keeping kernel sizes and strides coherent for both branches. 

\begin{figure}[t]
\begin{center}
\includegraphics[width=0.948\linewidth]{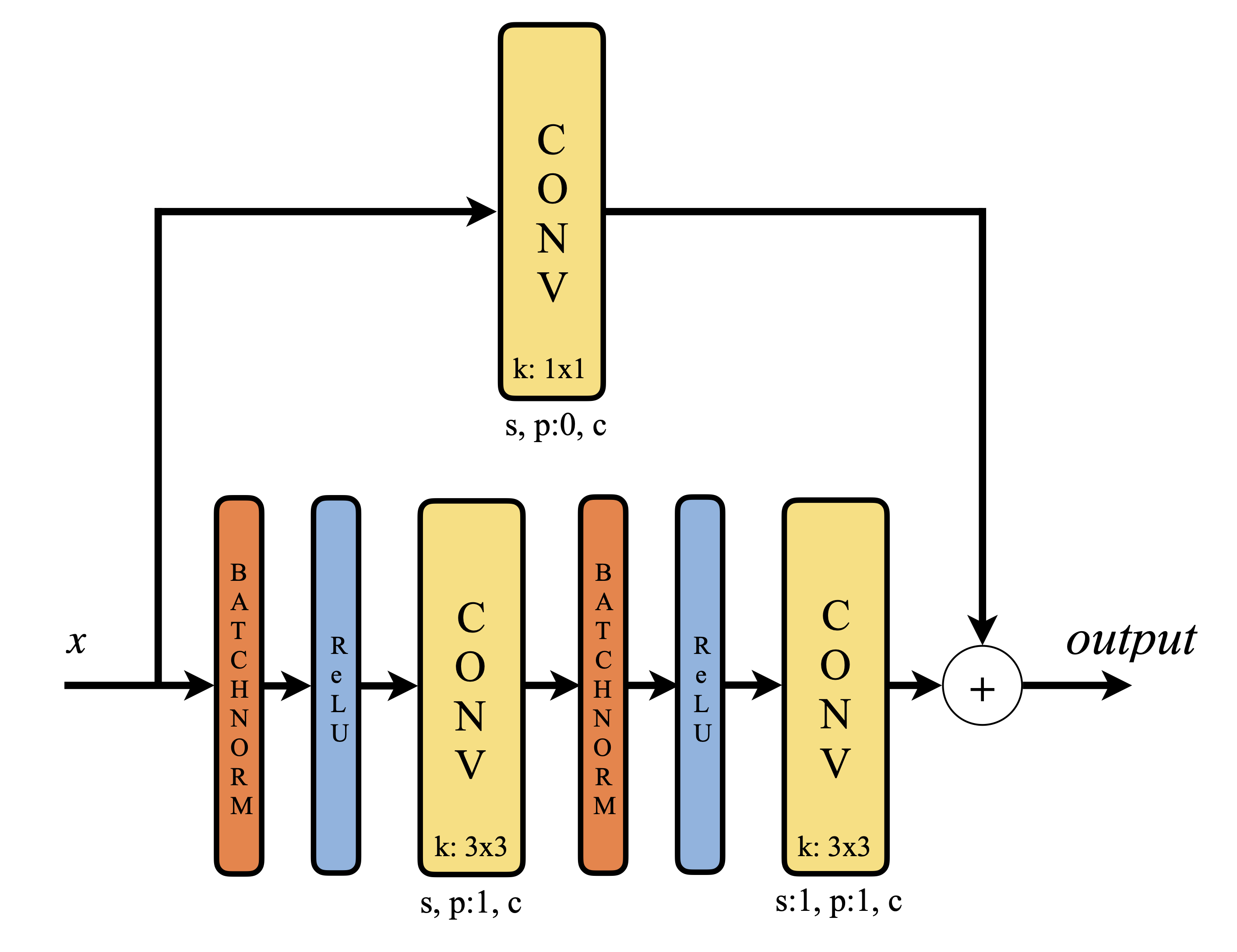}
\end{center}

 \caption{Illustration of a single residual block used in our design. Stride $s$ of the residual block is applied in the first convolutional layer and in the skip connection. The number of channels $c$ of the residual block is applied on each convolutional layer with same padding $p$.}
\label{fig:resblock}

\end{figure}

 \textit{Pose branch} consists of two residual blocks followed by a \textit{1x1} convolutional layer and a batch normalization. Residual blocks are designed as the same blocks that learn the imaginary parts in Deep Complex Networks \cite{trabelsi2017deepComplex}. The detailed view of a single residual block is given in Figure \ref{fig:resblock}. While both residual blocks have \textit{3x3} kernels, the first one has 32 channels with stride 1, whereas the second has 64 channels with stride 2. The following \textit{1x1} convolutional layer increases the dimensionality in feature space in order to match the dimensionality of Primary capsules. Since capsule poses are represented as pure quaternions requiring 3 imaginary parts, this layer has $\textit{N}\times 3$ channels where \textit{N} refers to the number of capsules in primary layer, and is set to 96. Then, batch normalization \cite{batchNorm} is applied to the output before forming Primary capsules. 
 
\textit{Activation branch} is relatively shallower as assumed that the existence probability can be simply extracted. Therefore, only one residual block is employed before the same pipeline in \textit{Pose branch} with \textit{N} channels. To ensure that the spatial alignment with \textit{Pose branch} is preserved, the residual block has \textit{3x3} kernel with stride 2. The number of channels in this block is set to 32.

Pose and activation outputs corresponding to the same receptive fields are grouped to compose Primary capsules $\left[u_i, a_i \right]$. Thus, each Primary capsule refers to the pose and activation of a visual entity in particular receptive field. Consecutive 3 convolutional capsule layers have 16 capsules and 1 stride with \textit{5x5} kernels. The last convolutional layer is fully-connected to Class capsule layer where the number of capsules are dependent on the number of classes in the dataset (\textit{e.g} for MNIST and smallNORB, the number of capsules are 10 and 5, respectively). All capsule layers are connected by EM Routing with 2 routing iterations, and the spread loss \eqref{eqn:Spread} is used without any manipulation on the original version in \cite{hinton2018matrix}.

\begin{equation}
    \label{eqn:Spread}
    L=\sum_{i \neq t }(max(0,m-(a_{ t }-a_{ i })))^{ 2 },\quad L = \sum_{i \neq t}L_i
\end{equation}



\section{Experiments}

As the main goal of QCN to achieve better generalization to the novel viewpoints, the experiments on smallNORB are conducted on a setup where training is made on a limited range of viewpoints. Additionally, benchmarking experiments are conducted on smallNORB, MNIST, FashionMNIST, SVHN and CIFAR10 datasets in order to show that QCN can achieve on-par performances with Matrix Capsules on multiple datasets. These datasets are chosen with respect to the presented Capsule Network results in the literature \cite{DynRouteCaps, hinton2018matrix,VariationalCaps}. Benchmarking results also contain the results of our implementation and the open-source IBM implementation \cite{gritzman2019IBMCaps}. In our design, the architecture and hyper-parameters (\textit{e.g} optimizer, initial learning rate, learning rate scheduler, etc.) are kept the same with our implementation of Matrix Capsules for all experiments. The only factor that may change is the batch size depending on the memory requirements of the datasets. 

In this study, even though the experimental results show that QCN achieves the state-of-the-art performance in case of generalization to novel viewpoints, the main purpose is to make a proof-of-concept design that quaternions form a compact way of representing the rotations and orientations for capsules rather than achieving the state-of-the-art performances in Capsule Network literature. Therefore, QCNs are not fine-tuned for any of the datasets, instead, the configuration that achieve the best performance on generalization to novel viewpoints setup is used.

\subsection{Implementation Details}
The implementation details regarding QCNs are explained in this section. First, a quaternion $q = q_0 + q_1 i + q_2 j + q_3 k$ is isomorphic to real-valued matrices ${^{R}{Q}}$ and ${^{L}{Q}}$ in case of quaternion product from right and left, as can be seen in \eqref{eqn:isoQr}.

\newcommand\scalemath[2]{\scalebox{#1}{\mbox{\ensuremath{\displaystyle #2}}}}

\begin{equation}
    \scalemath{0.83}{
    \label{eqn:isoQr}
    {^{R}{Q}}=\left[ \begin{matrix} q_0 &   -q_1    &   -q_2    & -q_3 \\ 
                                    q_1 &   q_0     &   -q_3    & q_2 \\ 
                                    q_2 &   q_3     &   q_0     & -q_1 \\
                                    q_3 &   -q_2    &   q_1     & q_0
    \end{matrix} \right], {^{L}{Q}}=\left[ \begin{matrix} q_0 &   -q_1    &   -q_2    &   -q_3 \\ 
                                    q_1 &   q_0     &   q_3     &   -q_2 \\ 
                                    q_2 &   -q_3    &   q_0     &   q_1 \\
                                    q_3 &   q_2     &   -q_1    &   q_0
    \end{matrix} \right]
    }
\end{equation}


Given these embeddings, vote calculation with quaternion rotation in \eqref{eqn:QuatVote} can be performed in the form of matrix multiplication as given in \eqref{eqn:MatFormQuatMult}, where ${^{L}{W}^{*}_{ij}}$ represent the quaternion rotor conjugate embedded to product from right matrix, and ${^{R}{W}_{ij}}$ represent quaternion rotor embedded to product from left matrix.

\begin{equation}
    \label{eqn:MatFormQuatMult}
      v_{ij} = {^{L}{W}^{*}_{ij}} {^{R}{W}_{ij}} u_{i} 
\end{equation}

\renewcommand{\arraystretch}{1.35}

The residual blocks in the activation and pose branches are initialized with uniform Kaiming distribution \cite{he2015delving}, while \textit{1x1} convolutional layers that follow the residual blocks in both branches are initialized with uniform Xavier distribution \cite{glorot2010understanding}. Each element of the rotation axes $\Bar{w}_{ij}$ in quaternion capsule layers are initialized from $U(-1, 1)$. Initialization of the transformation matrices to identity with noise on off-diagonals \cite{VariationalCaps} proves to stabilize the training by restricting the amount of transformation in the earlier stages. This is analogous to the idea of restricting the initial rotations in QCNs to a certain range. Therefore,  $\theta_{ij}$ is initialized by the uniform distribution in the range of $\left[-\pi, \pi \right]$ as in \cite{parcollet2018quaternionRNN}. Another important detail is the $m$ scheduling in Spread loss. As stated in \cite{gritzman2019IBMCaps}, $m$ is updated by \eqref{eqn:mschedule} until $m$ reaches $0.9$, where $\sigma$ is the sigmoid function.

\begin{equation}
    \label{eqn:mschedule}
        m = 0.2 + 0.79 * \sigma\left( min\left( 10, \frac{step}{50000} - 4  \right) \right)
\end{equation}

\begin{table}[t]
\caption{Test error rate comparison of the reported results on the original paper (EM), Capsule Routing via Variational Bayes (VB), our implementation (EM*) of Matrix Capsules, proposed Quaternion Capsules (QCN) and CNN results in \cite{hinton2018matrix} on novel viewpoint setup (\textit{i.e.} azimuth and elevation).}
\center{
\scalebox{1.2}{
\begin{tabular}{l|ccccc}
\hline \hline
\textbf{Viewpoints} & \multicolumn{5}{c}{\textbf{Azimuth} (\%)}  \\ 
(Models)    & QCN   & EM*   & VB  & EM  & CNN   \\ \hline
Novel       & \textbf{7.5}  & 13.4  & 11.3  & 13.5   & 20.0  \\ 
Familiar    & 3.7           & 3.7   & 3.7   & 3.7    & 3.7  \\ \hline
\textbf{Viewpoints} & \multicolumn{5}{c}{\textbf{Elevation} (\%)} \\
(Models)       & QCN   & EM*   & VB & EM  & CNN  \\ \hline
Novel       & \textbf{11.5} & 15.8  & 11.6  & 12.3  & 17.8  \\ 
Familiar    & 4.4           & 4.0   & 4.3   & 4.3   &  4.3 \\ \hline \hline
\end{tabular}}
}
\label{tab:NovelViewResults}
\end{table}

\subsection{Generalization to Novel Viewpoints}

The experiments of generalization to novel viewpoints are conducted on smallNORB dataset with given setup in \cite{hinton2018matrix}. The smallNORB \cite{lecun2004learning} is a 3D object recognition dataset which consists of $96 \times 96$ stereo image samples of 5 classes under 6 lighting conditions, 9 elevations (\textit{i.e.} every 5 degrees from 30 to 70 degrees), and 18 azimuths (\textit{i.e.} every 20 degrees from 0 to 340). In this setup, there are two different cases in which the dataset is reserved with respect to azimuth angles and elevations, instead of instances. For azimuth experiments, the azimuth angles of (300, 320, 340, 0, 20, 40) are used for training, and the tests are made on the remaining azimuth angles. Secondly, QCN is trained on 3 smaller elevations (30, 35, 40 degrees from the horizontal), and tested on the remaining 6 larger elevations (45, 50, 55, 60, 65, 70 degrees from the horizontal). 

For the sake of fair comparison, the test set (novel viewpoints) accuracy is measured once the training set (familiar viewpoints) accuracy is matched with the reported values in \cite{hinton2018matrix}. QCN outperforms both Matrix Capsules with EM Routing (EM) and Capsule Routing via Variational Bayes (VB) with a significant performance improvement for both azimuth and elevation setups, when compared to VB and EM, respectively.
The baseline CNN results reported in \cite{hinton2018matrix} are outperformed by all of the capsule architectures by a large margin. This baseline architecture contains two convolutional layers with 32 and 64 channels followed by a fully-connected layer with 1024 neurons, and summing up to 4.2M parameters. 

\begin{table*}[t]
\center

\caption{Comparison of QCN test error rates with the reported results of  Matrix Capsules (EM), IBM's (EM-IBM) and our (EM*) implementations of Matrix Capsules, and Capsule Routing via Variational Bayes (VB). $-$: Not reported, $\ddagger$: We run their open-source code on corresponding dataset with their default hyper-parameters.}
\resizebox{\linewidth}{!}{%
\begin{tabular}{l|cc|cc|cc|cc|cc}
\hline \hline  
\multirow{2}{*}{\textbf{Models}} & \multicolumn{2}{c|}{\textbf{smallNORB}} & \multicolumn{2}{c|}{\textbf{MNIST}} & \multicolumn{2}{c|}{\textbf{FashionMNIST}} & \multicolumn{2}{c|}{\textbf{SHVN}} & \multicolumn{2}{c}{\textbf{CIFAR-10}} \\  
                        & Error (\%)        & \# of Params        & Error (\%)      & \# of Params      & Error (\%)          &
                        \# of Params         & Error (\%)      & \# of Params     & Error (\%)        & \# of Params       \\   \hline  
EM \cite{hinton2018matrix}    &      1.8             &    {\raise.17ex\hbox{$\scriptstyle\sim$}}310K                 &   0.44              &   --
&      --               &      --                &      --           &         --         &        11.9        &           --      \\ 
EM-IBM  \cite{gritzman2019IBMCaps}    &      4.6              &       {\raise.17ex\hbox{$\scriptstyle\sim$}}335K              &     1.23\textsuperscript{$\ddagger$}            & {\raise.17ex\hbox{$\scriptstyle\sim$}}337K\textsuperscript{$\ddagger$}
&       10.44\textsuperscript{$\ddagger$}             &        {\raise.17ex\hbox{$\scriptstyle\sim$}}337K\textsuperscript{$\ddagger$}            &         --        &        --          &          --         &         --           \\ 
VB  \cite{VariationalCaps}   &         1.6          &        {\raise.17ex\hbox{$\scriptstyle\sim$}}169K             &        --         & --
&        5.2             &         {\raise.17ex\hbox{$\scriptstyle\sim$}}172K             &       3.9          &       {\raise.17ex\hbox{$\scriptstyle\sim$}}323K           &        11.2            &          {\raise.17ex\hbox{$\scriptstyle\sim$}}323K          \\ 
EM*     &         3.40          &   {\raise.17ex\hbox{$\scriptstyle\sim$}}317K            &       0.89          &          {\raise.17ex\hbox{$\scriptstyle\sim$}}319K         & 9.74
&      {\raise.17ex\hbox{$\scriptstyle\sim$}}319K         &       8.19         &      {\raise.17ex\hbox{$\scriptstyle\sim$}}320K         &     17.76              &     {\raise.17ex\hbox{$\scriptstyle\sim$}}460K               \\  \hline 
QCN                     &       2.29            &        {\raise.17ex\hbox{$\scriptstyle\sim$}}188K             &        0.37         &       {\raise.17ex\hbox{$\scriptstyle\sim$}}187K            &         6.92           & {\raise.17ex\hbox{$\scriptstyle\sim$}}187K
&      4.63        &        {\raise.17ex\hbox{$\scriptstyle\sim$}}189K         &         13.92          &         {\raise.17ex\hbox{$\scriptstyle\sim$}}189K           \\  \hline \hline
\end{tabular}%
}

\label{tab:GeneralResults}
\end{table*}

\subsection{Benchmarking Results}

\noindent
\textbf{smallNORB:}
In the default settings of smallNORB, training and test sets contain 23,400 samples where each class has 10 different instances. The original format of the dataset has a standard classification setup where test set instances are not seen during training. During training, samples are initially resized to $48\times48$, and $32\times32$ random patches of binocular images fed to the network with a batch size of 64. Center cropping is applied to the test images. QCN achieves 2.3\% test error rate with almost half of the parameters as in Matrix Capsules. Even though QCN surpasses IBM's and our implementations of Matrix Capsules, it falls behind of the accuracy reported in the official paper by 0.6\%. \newline

\noindent
\textbf{MNIST:}
Experiments on MNIST are conducted on the standard setup without any modification. Even though QCN specializes on the rotations, it can also achieve on-par classification performance with Matrix Capsules on MNIST. QCN achieves 0.37\% test error rate surpassing both performances in our implementation and as reported in \cite{hinton2018matrix}, which are 0.89\% and 0.44\%. Note that MNIST is a toy problem, rather than being a distinctive in terms of test performances. However, the fact that QCN surpasses Matrix Capsules, indicates that quaternions have at least the same capability to achieve such performance with less number of parameters and more compact representation. \newline

\noindent
\textbf{FashionMNIST:}
FashionMNIST can be considered as a more difficult dataset than the previous ones whose 
the properties are the same with MNIST, except the number of samples in each split (\textit{i.e.} 60,000 training and 10,000 test samples with the size of $28\times28$). During training of QCN and our Matrix Capsules implementation, any augmentation technique is applied to the samples. As a result, QCN, with a relatively shallow network, achieves 6.92\% test error rate, which is slightly worse than VB that achieves 5.2\%. On the other hand, our implementation of Matrix Capsules achieve only 9.74\% test error rate falling behind of QCN with a wide margin. \newline

\noindent
\textbf{SVHN:}
SVHN consists of RGB real-world house number images with 10 classes for each digit. Experiments are conducted on 76,257 samples, and tested on 26,032 samples. On both sets, the samples are scaled to $32\times32$, and normalized before feeding to the network. While QCN achieves 4.63\% test error rate, VB and our implementation of Matrix Capsules achieve 3.9\% and 8.19\% test error rates, respectively. \newline

\noindent
\textbf{CIFAR-10:}
CIFAR-10 consists of 6000 RGB images per each of 10 classes, where training and test sets have 50,000 and 10,000 samples with the size of $32\times32$. Training samples are zero-padded by 4 pixels, and randomly cropped to $32\times32$ patches. Lastly, horizontal flipping is applied to all training samples before feeding them to the network. Note that test samples are not modified. QCN achieves 13.9\% test error rate while \cite{hinton2018matrix} reported 11.9\% test error rate, which is achieved by increasing the number of neurons in the hidden layer to 256. However, in our main design, the architecture is identical for all datasets. Additionally, our implementation of Matrix Capsules with 256 neurons in hidden layer achieve 17.76\% test error rate. 
\\

\begin{table}[]
\caption{The effect of branching on the error rates and parameters for both Matrix and Quaternion Capsules. Non-branched versions of EM\textsuperscript{$\ddagger$} and QCN have two consecutive residual blocks with 64 and 96 channels until Primary capsules to ensure Primary capsules have the same number of input channels in both for a fair comparison.}
\scalebox{1.1}{
\begin{tabular}{l|ccc|ccc} \hline \hline
\multicolumn{1}{c|}{\multirow{3}{*}{\textbf{Models}}} & \multicolumn{3}{c|}{\textbf{Error Rate (\%)}} & 
\multicolumn{3}{c}{\textbf{{\raise.17ex\hbox{$\scriptstyle\sim$}}\# Parameters}} \\ 
\multicolumn{1}{c|}{}                                 & with                  & without         & diff.       & with                  & without    &  diff.              \\ \hline
EM\textsuperscript{$\ddagger$}                                                   & 3.66        &           3.44               &  0.22       &        411K    & 533K   & -122K               \\ 
QCN  &        2.29   & 3.72  &   \textbf{-1.43}   & 188K  & 298K & -110K    \\               \hline \hline
\end{tabular}}
\end{table}

\noindent
\textbf{Effect of branching:}
Since we have used the complex component extractor as in \cite{trabelsi2017deepComplex}, it is important to determine its effects on the performance of capsules. Therefore, we have conducted additional experiments on smallNORB with branched and non-branched versions of QCN and Matrix Capsules. Note that the branched version of Matrix Capsules is consist of the same capsule extractor layers as in QCN, while non-branched versions have two consecutive residual blocks with 64 and 96 channels, as shown in Figure \ref{fig:resblock}.
The results demonstrate that using separate branches for pose and activation increases the overall performance of QCN, though the number of parameters is reduced. For QCN, this is expected as we use the blocks that are empirically proven to be  useful for extracting the complex components \cite{trabelsi2017deepComplex}. On the other hand, EM\textsuperscript{$\ddagger$} yields slightly larger error rate when branching is applied.

\section{Conclusion}
In this paper, we propose QCNs that represent capsule orientations and their transformations with quaternions. Representing the rotation mappings between the parent and child capsules with quaternions dramatically increases the generalization performance in novel viewpoints setup, and outperforms the performances of the previous Capsule Networks. Moreover, using quaternions for such operations, instead of matrices, reduces the number of parameters in the network by half, and it can still achieve on-par performances on several well-known benchmarking datasets. With this promising results, QCNs can provide a novel research direction to capsule representations that affects on both the computational efficiency and the generalization performance. Since common deep learning tools do not have native support for the quaternion operations, we cannot fully-benefit from the computational efficiency of quaternion rotations for the time being.  Research on initialization, quaternion-specific objective functions, and creating an efficient quaternion operation support for PyTorch are left for the future work. 

\section{Acknowledgement}

We would like to thank Prof. Erhan Öztop for great insights and discussions on Quaternion Capsules, and Doğa Yılmaz for his contributions to the writing process.

\bibliographystyle{IEEEtran}
\bibliography{conf}

\begin{thebibliography}{10}
\providecommand{\url}[1]{#1}
\csname url@samestyle\endcsname
\providecommand{\newblock}{\relax}
\providecommand{\bibinfo}[2]{#2}
\providecommand{\BIBentrySTDinterwordspacing}{\spaceskip=0pt\relax}
\providecommand{\BIBentryALTinterwordstretchfactor}{4}
\providecommand{\BIBentryALTinterwordspacing}{\spaceskip=\fontdimen2\font plus
\BIBentryALTinterwordstretchfactor\fontdimen3\font minus
  \fontdimen4\font\relax}
\providecommand{\BIBforeignlanguage}[2]{{%
\expandafter\ifx\csname l@#1\endcsname\relax
\typeout{** WARNING: IEEEtran.bst: No hyphenation pattern has been}%
\typeout{** loaded for the language `#1'. Using the pattern for}%
\typeout{** the default language instead.}%
\else
\language=\csname l@#1\endcsname
\fi
#2}}
\providecommand{\BIBdecl}{\relax}
\BIBdecl

\bibitem{krizhevsky2012imagenet}
A.~Krizhevsky, I.~Sutskever, and G.~E. Hinton, ``Image{N}et {C}lassification
  with {D}eep {C}onvolutional {N}eural {N}etworks,'' in \emph{Advances in
  Neural Information Processing Systems}, 2012, pp. 1097--1105.

\bibitem{ren2015faster}
S.~Ren, K.~He, R.~Girshick, and J.~Sun, ``Faster {R-CNN}: Towards {R}eal-{T}ime
  {O}bject {D}etection with {R}egion {P}roposal {N}etworks,'' in \emph{Advances
  in {N}eural {I}nformation {P}rocessing {S}ystems}, 2015, pp. 91--99.

\bibitem{he2016residual}
K.~He, X.~Zhang, S.~Ren, and J.~Sun, ``{D}eep {R}esidual {L}earning for {I}mage
  {R}ecognition,'' in \emph{The IEEE Conference on Computer Vision and Pattern
  Recognition (CVPR)}, June 2016.

\bibitem{Redmon_2016_YOLO}
J.~Redmon, S.~Divvala, R.~Girshick, and A.~Farhadi, ``{You Only Look Once:
  Unified, Real-Time Object Detection},'' in \emph{The IEEE Conference on
  Computer Vision and Pattern Recognition (CVPR)}, June 2016.

\bibitem{DeeperConv}
C.~Szegedy, W.~Liu, Y.~Jia, P.~Sermanet, S.~Reed, D.~Anguelov, D.~Erhan,
  V.~Vanhoucke, and A.~Rabinovich, ``{Going Deeper With Convolutions},'' in
  \emph{The IEEE Conference on Computer Vision and Pattern Recognition (CVPR)},
  June 2015.

\bibitem{DynRouteCaps}
S.~Sabour, N.~Frosst, and G.~E. Hinton, ``{Dynamic Routing Between Capsules},''
  in \emph{Advances in {N}eural {I}nformation {P}rocessing {S}ystems}, 2017,
  pp. 3856--3866.

\bibitem{hinton2018matrix}
G.~E. Hinton, N.~Frosst, and S.~Sabour, ``{Matrix capsules with EM routing},''
  in \emph{{International Conference on Learning Representations (ICLR)}},
  2018.

\bibitem{VariationalCaps}
F.~D.~S. Ribeiro, G.~Leontidis, and S.~Kollias, ``{Capsule Routing via
  Variational Bayes},'' \emph{arXiv preprint arXiv:1905.11455}, 2019.

\bibitem{kosiorek2019stacked}
\BIBentryALTinterwordspacing
A.~R. Kosiorek, S.~Sabour, Y.~W. Teh, and G.~E. Hinton, ``Stacked {C}apsule
  {A}utoencoders,'' in \emph{Advances in Neural Information Processing
  Systems}, 2019. [Online]. Available: \url{https://arxiv.org/abs/1906.06818}
\BIBentrySTDinterwordspacing

\bibitem{grassia1998practical}
F.~S. Grassia, ``Practical parameterization of rotations using the exponential
  map,'' \emph{Journal of graphics tools}, vol.~3, no.~3, pp. 29--48, 1998.

\bibitem{pavllo2018quaternet}
D.~Pavllo, D.~Grangier, and M.~Auli, ``Quaternet: A quaternion-based recurrent
  model for human motion,'' \emph{arXiv preprint arXiv:1805.06485}, 2018.

\bibitem{matsui2004quaternion}
N.~Matsui, T.~Isokawa, H.~Kusamichi, F.~Peper, and H.~Nishimura, ``Quaternion
  neural network with geometrical operators,'' \emph{Journal of Intelligent \&
  Fuzzy Systems}, vol.~15, no. 3, 4, pp. 149--164, 2004.

\bibitem{trabelsi2017deepComplex}
C.~Trabelsi, O.~Bilaniuk, Y.~Zhang, D.~Serdyuk, S.~Subramanian, J.~F. Santos,
  S.~Mehri, N.~Rostamzadeh, Y.~Bengio, and C.~J. Pal, ``Deep {C}omplex
  {N}etworks,'' \emph{arXiv preprint arXiv:1705.09792}, 2017.

\bibitem{gaudet2018deepQuat}
C.~J. Gaudet and A.~S. Maida, ``Deep {Q}uaternion {N}etworks,'' in \emph{2018
  International Joint Conference on Neural Networks (IJCNN)}.\hskip 1em plus
  0.5em minus 0.4em\relax IEEE, 2018, pp. 1--8.

\bibitem{parcollet2018quaternionRNN}
T.~Parcollet, M.~Ravanelli, M.~Morchid, G.~Linar{\`e}s, C.~Trabelsi,
  R.~De~Mori, and Y.~Bengio, ``Quaternion {R}ecurrent {N}eural {N}etworks,'' in
  \emph{International Conference on Learning Representations (ICLR)}, 2019.

\bibitem{hinton2011transforming}
G.~E. Hinton, A.~Krizhevsky, and S.~D. Wang, ``{Transforming Auto-encoders},''
  in \emph{Proceedings of the 21th International Conference on Artificial
  Neural Networks}, ser. ICANN'11, 2011, pp. 44--51.

\bibitem{lecun2004learning}
Y.~Lecun, F.~Huang, and L.~Bottou, ``{Learning Methods for Generic Object
  Recognition with Invariance to Pose and Lighting},'' in \emph{IEEE Conference
  on Computer Vision and Pattern Recognition (CVPR)}, 01 2004, pp. II--97.

\bibitem{kurakin2016adversarial}
A.~Kurakin, I.~Goodfellow, and S.~Bengio, ``Adversarial examples in the
  physical world,'' \emph{arXiv preprint arXiv:1607.02533}, 2016.

\bibitem{brendel2017comment}
W.~Brendel and M.~Bethge, ``Comment on "{B}iologically inspired protection of
  deep networks from adversarial attacks",'' \emph{arXiv preprint
  arXiv:1704.01547}, 2017.

\bibitem{Li2018NeuralNE}
H.~Li, X.~Guo, B.~Dai, O.~Wanli, and X.~Wang, ``{Neural Network
  Encapsulation},'' in \emph{The European Conference on Computer Vision
  (ECCV)}, September 2018.

\bibitem{lenssen2018groupCaps}
J.~E. Lenssen, M.~Fey, and P.~Libuschewski, ``Group {E}quivariant {C}apsule
  {N}etworks,'' in \emph{Advances in Neural Information Processing Systems},
  2018, pp. 8844--8853.

\bibitem{cohen2016group}
T.~S. Cohen and M.~Welling, ``{Group Equivariant Convolutional Networks},'' in
  \emph{Proceedings of the 33rd International Conference on International
  Conference on Machine Learning - Volume 48}, ser. ICML'16, 2016, pp.
  2990--2999.

\bibitem{duarte2018VideoCapsNet}
K.~Duarte, Y.~Rawat, and M.~Shah, ``{VideoCapsuleNet: A Simplified Network for
  Action Detection},'' in \emph{Advances in Neural Information Processing
  Systems 31}, 2018, pp. 7610--7619.

\bibitem{lalonde2018SegCaps}
R.~LaLonde and U.~Bagci, ``{Capsules for Object Segmentation},'' \emph{arXiv
  preprint arXiv:1804.04241}, 2018.

\bibitem{kinli2019RCCapsNet}
F.~Kinli, B.~Ozcan, and F.~Kirac, ``{Fashion Image Retrieval with Capsule
  Networks},'' in \emph{The IEEE International Conference on Computer Vision
  (ICCV) Workshops}, Oct 2019.

\bibitem{hirose2004complex}
A.~Hirose, ``{Complex-Valued Neural Networks: Theories and Applications (Series
  on Innovative Intelligence, 5)},'' 2004.

\bibitem{zimmermann2011comparison}
H.~G. Zimmermann, A.~Minin, and V.~Kusherbaeva, ``{Comparison of the Complex
  Valued and Real Valued Neural Networks Trained with Gradient Descent and
  Random Search Algorithms},'' in \emph{Proc. of ESANN 2011}, 2011.

\bibitem{danihelka2016associative}
I.~Danihelka, G.~Wayne, B.~Uria, N.~Kalchbrenner, and A.~Graves, ``{Associative
  Long Short-Term Memory},'' in \emph{Proceedings of The 33rd International
  Conference on Machine Learning}, vol.~48, 2016, pp. 1986--1994.

\bibitem{arjovsky2016unitary}
M.~Arjovsky, A.~Shah, and Y.~Bengio, ``{Unitary Evolution Recurrent Neural
  Networks},'' in \emph{Proceedings of the 33rd International Conference on
  International Conference on Machine Learning - Volume 48}, ser. ICML'16,
  2016, pp. 1120--1128.

\bibitem{nitta2002critical}
T.~Nitta, ``On the critical points of the complex-valued neural network,'' in
  \emph{Proceedings of the 9th International Conference on Neural Information
  Processing, 2002. ICONIP'02.}, vol.~3.\hskip 1em plus 0.5em minus 0.4em\relax
  IEEE, 2002, pp. 1099--1103.

\bibitem{hirose2012complexgeneralization}
A.~Hirose and S.~Yoshida, ``{Generalization Characteristics of Complex-Valued
  Feedforward Neural Networks in Relation to Signal Coherence},'' \emph{IEEE
  Transactions on Neural Networks and Learning Systems}, vol.~23, pp. 541--551,
  2012.

\bibitem{zhu2018quaternionconv}
X.~Zhu, Y.~Xu, H.~Xu, and C.~Chen, ``{Quaternion Convolutional Neural
  Networks},'' in \emph{The European Conference on Computer Vision (ECCV)},
  September 2018.

\bibitem{parcollet2018quaternionEndtoEnd}
T.~Parcollet, Y.~Zhang, M.~Morchid, C.~Trabelsi, G.~Linarès, R.~De~Mori, and
  Y.~Bengio, ``{Quaternion Convolutional Neural Networks for End-to-End
  Automatic Speech Recognition},'' 06 2018.

\bibitem{pletincks1988useQuat}
D.~Pletincks, ``The use of quaternions for animation, modelling and
  rendering,'' in \emph{New Trends in Computer Graphics}.\hskip 1em plus 0.5em
  minus 0.4em\relax Springer, 1988, pp. 44--53.

\bibitem{salimans2016weight}
T.~Salimans and D.~P. Kingma, ``{Weight Normalization: A Simple
  Reparameterization to Accelerate Training of Deep Neural Networks},'' in
  \emph{Advances in Neural Information Processing Systems 29}, 2016, pp.
  901--909.

\bibitem{batchNorm}
S.~Ioffe and C.~Szegedy, ``Batch normalization: Accelerating deep network
  training by reducing internal covariate shift,'' in \emph{Proceedings of the
  32nd International Conference on Machine Learning}, vol.~37, 07--09 Jul 2015,
  pp. 448--456.

\bibitem{gritzman2019IBMCaps}
A.~D. Gritzman, ``{Avoiding Implementation Pitfalls of ``Matrix Capsules with
  EM Routing'' by Hinton et al.}'' in \emph{Human Brain and Artificial
  Intelligence}.\hskip 1em plus 0.5em minus 0.4em\relax Singapore: Springer
  Singapore, 2019, pp. 224--234.

\bibitem{he2015delving}
K.~He, X.~Zhang, S.~Ren, and J.~Sun, ``{Delving Deep into Rectifiers:
  Surpassing Human-Level Performance on ImageNet Classification},'' in
  \emph{Proceedings of the IEEE International Conference on Computer Vision
  (ICCV)}, 2015, pp. 1026--1034.

\bibitem{glorot2010understanding}
X.~Glorot and Y.~Bengio, ``Understanding the difficulty of training deep
  feedforward neural networks,'' in \emph{Proceedings of the Thirteenth
  International Conference on Artificial Intelligence and Statistics}, vol.~9,
  13--15 May 2010, pp. 249--256.

\end{thebibliography}

\end{document}